\documentclass{article}
\usepackage{amsmath,amssymb,amsfonts}
\usepackage{algorithmic}
\usepackage{graphicx}
\usepackage{textcomp}

\usepackage{pdflscape}
\usepackage{afterpage}
\usepackage{capt-of}

\usepackage{rotating}
\usepackage{comment}

\usepackage[sorting=none]{biblatex}
\usepackage{booktabs}
\usepackage{multirow}

\def\BibTeX{{\rm B\kern-.05em{\sc i\kern-.025em b}\kern-.08em
    T\kern-.1667em\lower.7ex\hbox{E}\kern-.125emX}}

\DeclareTextFontCommand{\textbf}{\bfseries}
\DeclareTextFontCommand{\textit}{\itshape}

\begin{document}

{\Large \bf A Systematic Review of Available Datasets in Additive Manufacturing\footnote{Acknowledgement: This work was funded by Science Foundation Ireland through the I-FORM the SFI Research Centre for Advanced Manufacturing (16/RC/3872) and Insight the SFI Research Centre for Data Analytics (SFI/12/RC/2289\_P2).}}

\vspace{1cm}

{\large \bf Xiao Liu$^{(1,2)}$, Alessandra Mileo$^{(1,2,3)}$ and Alan F. Smeaton$^{(2,3)}$}

\vspace{1cm}

\noindent 
{[1] I-FORM SFI Centre for Advanced Manufacturing, Dublin City University, Glasnevin, Dublin 9, Ireland.}\\
{[2] School of Computing, Dublin City University, Glasnevin, Dublin 9, Ireland.}\\
{[3] Insight SFI Centre for Data Analytics, Dublin City University, Glasnevin, Dublin 9, Ireland.}\\


\begin{abstract}

In-situ monitoring incorporating data from visual and other sensor technologies, allows the collection of extensive datasets during the Additive Manufacturing (AM) process. These datasets have potential for determining the quality of the manufactured output and the detection of defects through the use of Machine Learning during the manufacturing process.  Open and annotated datasets derived from AM processes are necessary for the machine learning community to address this opportunity, which creates difficulties in the application of computer vision-related machine learning in AM. This systematic review  investigates the availability of open image-based datasets originating from AM processes that align with a number of pre-defined selection criteria. The review identifies existing gaps among the current image-based datasets in the domain of AM, and points to the need for greater availability of open datasets in order to allow quality assessment and defect detection during additive manufacturing, to develop.

\end{abstract}




\section{Introduction}
\label{sec:introduction}

{Large} and open-source datasets of annotated images containing up to millions of training examples such as ImageNet \cite{deng2009imagenet}, COCO (Common Objects in Context) \cite{lin2014microsoft} and Pascal VOC \cite{everingham2010pascal} are readily available to machine learning and deep learning researchers. ImageNet for example contains more than 14 million annotated images and the datasets have allowed machine learning to develop hugely over recent years, especially because the datasets are open, easily available and re-used by many researchers. These datasets are usually not specific to any single domain, they are general purpose though there are also many examples of annotated image datasets which are in specific domains.  Examples of domain-specific datasets are in areas like medical imaging with the Cancer Imaging Archive (TCIA) \cite{clark2013cancer}, the Breast Cancer Digital Repository (BCDR) and Medical Imaging Multimodality Breast Cancer Diagnosis User Interface (MIMBCD-UI) \cite{lopez2012bcdr} which are images of cancer that have been used extensively.  Another example is in the area of recognition of birds where the Caltech-UCSD Birds-200-2011dataset \cite{wah2011caltech} contains images of 200 different bird species.
 
In the area of additive manufacturing (AM), also known as 3D printing, we have seen a huge increase in the amount of  in-situ monitoring during the actual manufacturing process \cite{mccann2021situ}. This includes the use of process sensors recording a wide range of optical, acoustic and thermal signals which gather data on the melt-pool as layers are deposited in creating the manufactured output. The aim of such process monitoring is to measure quality and to detect potential microstructure defects in the output such as internal porosity variance or cracking as the part is being manufactured. Ultimately this means that manufacturing that part can stop and there is no further wastage of resources, or perhaps subsequent deposited layers can compensate for the recognised defect, and manufacturing can continue.

Given the rapid development in data collocation in the domain of additive manufacturing, it is reasonable to  examine the availability of datasets of annotated images taken from the AM process where such annotated images can be used to train machine learning algorithms to recognise microstructure defects. 

Acquiring annotated process monitoring data is cost-prohibitive in the AM area as shown by Manan and Shao \cite{mehta2022federated}. Several very recent (2020-2022) survey and review papers on the topic of image datasets \cite{yang2022survey} \cite{ouali2020overview} \cite{van2020survey} each clearly state that sample images from the AM process, labelled with annotations of microstructure defects in the manufacture,  are often difficult, expensive, and time-consuming to obtain. By contrast, unlabelled data can be easily or inexpensively obtained. 
Consequently, it is desirable to leverage a large number of unlabelled image data for improving the learning performance when this can be combined with a small number of labelled samples. Some researchers have called this the ``Small Data Challenges in Big Data Era” \cite{qi2020small}. 

In this article we present the results of a systematic review  to identify the gap between currently available image datasets from AM processes and what is available in other areas, usually a large number of standard open-source datasets which can be used in other applications of machine learning. We focus on images which can identify microstructure defects in the melt-pools such as porosity variance and cracks.

\section{Overview of this Systematic Review}

A systematic review of scientific literature  employs systematic and explicit methods to identify, select, and critically assess relevant primary research studies and involves the extraction and analysis of data from those studies. The methodologies utilized in a systematic review must be both reproducible and transparent to ensure that the outcomes  can be independently verified by other researchers. Typically, as outlined in  Wright et al. \cite{wright2007write}, a systematic review involves seven stages: (1) team formation and determining who completes what tasks, (2) questioning what is the underlying aim of the review, (3) planning the methods to be used for executing the data gathering, (4) searching through the gathered data and screening the results, (5) managing and reporting the screened output, (6) extracting data and trends from the screened data, aggregating and noting patterns and gaps and finally (7) drawing conclusions which tie in with the earlier stage of questioning the aim of the review. While some of these steps may not be directly applicable to the  context of this review such as team formation (1), we will  follow this sequence and describe the steps we have taken in the following sections.

As we set out to complete a systematic review of the literature on available datasets in AM we have some underlying questions which we want to investigate, the second stage known as ``questioning”. 
%
This leads to the following set of questions which will help guide the review
\begin{enumerate}
\item Is the current availability of annotated or unannotated image datasets from AM processes adequate for computer vision applications?
\item For  existing open AM image datasets, are they suitable or readily available to machine learning and deep learning researchers?
\item Are there many research programs and applications proposed for the development of image datasets that are specific to AM? 
\end{enumerate}

The selection criteria for inclusion is an important part of the third step of  systematic review mentioned earlier and covers the methods and strategies for carrying out the data gathering or in this case the search across multiple selected databases. These criteria specify the conditions and characteristics that published research studies with datasets must meet in order to be considered for inclusion in this review, ensuring that the review is systematic and objective.
\label{sec:criteria}
For inclusion of a dataset in the resulting selection, a dataset should meet the following 5 eligibility criteria:
\begin{enumerate}
\item Must be open and accessible;
\item Should be in the format of images and not only images of figures or charts since this is our focus and numerical datasets and videos will be excluded;
\item Must be related to the domain of additive manufacturing, not to other 3D-printed work;
\item Machine learning or deep learning (ML/DL) related applications must be involved in using the datasets which may result in different categorisations;
\item Should be readily available for ML/DL practices i.e. there should be no need for heavy computational pre-processing before putting them into use.
\end{enumerate}

\section{Database Searches for AM Datasets}

This section presents the strategy for the fourth step in the review process, namely
searching and gathering data from different sources, and screening the results.
A description of the search terms and query syntax for each of the databases where the searching has been carried out, along with the numbers of retrieved items for each search is presented as Appendix~A.

The nine databases chosen in the searching stage are DOE Data Explorer, Mendeley, Figshare, Zenodo, AmeriGEO, NIST, Kaggle, DataCite and Google Dataset Search. 
The  databases  and their corresponding URLs are shown in Table~\ref{tab:databases}

\begin{table}[ht]
\centering
\caption{Websites used in the searching stage}
\label{tab:databases}
\begin{tabular}{@{}cc@{}}
\toprule
Database Name         & URL                                        \\ \midrule
DOE Data Explorer     & \url{https://www.osti.gov/dataexplorer/}        \\
Mendeley              & \url{https://data.mendeley.com/}                 \\
Figshare              & \url{https://figshare.com/}                      \\
Zenodo                & \url{https://zenodo.org/}                        \\
AmeriGEO              & \url{https://data.amerigeoss.org/}               \\
NIST                  & \url{https://data.nist.gov/}                     \\
Kaggle                & \url{https://www.kaggle.com/}                    \\
DataCite              & \url{https://datacite.org/}                      \\
Google Dataset Search & \url{https://datasetsearch.research.google.com/} \\ \bottomrule
\end{tabular}%
\end{table}

Some of these databases like Figshare and Zenodo are standalone and store and index their own content while others like  Kaggle and Google Dataset Search are also aggregators meaning they store their own content but also index the content from other sources. This means that we can expect to find duplicate entries across the search results from individual sources, which the screening will detect.

Unlike search engines for bibliographic databases which have metadata fields like title, author, abstract, affiliations and other components, most search engines for datasets  do not have these structures. For this reason, the queries that were used will search for plain text that contains keywords such as  ``additive manufacturing” to specify the domain; ``image” to define the type of data and ``computer vision”, ``machine learning” and ``deep learning” to further indicate the desired applications. Boolean operators such as ‘AND’ are sometimes used to combine terms to form full queries and filter options have been applied in the relevant databases in order to limit the category and type of results. In some databases, basic searches may not be adequate to retrieve results in a reasonable range and the results usually contain too many non-relevant items. In such situations, advanced searches were conducted to obtain results with better precision. 

Initial searches were carried out on May 2th, 2023, and the dates of the search results are up to May 8th, 2023. The search stage is an interactive back and forth process that executes queries with different combinations of keywords on the selected databases. Results indicate the number of datasets which match the search criteria and depending on this number the searcher may broaden or narrow the search depending on whether too few or too many results are identified. This process continues until the search indicates a reasonable number of relevant results. Table~\ref{tab:search_results} shows the numbers of retrieved results from each database search.

\begin{table}[ht]
\centering
\caption{The numbers of results after searching on the databases}
\label{tab:search_results}
\begin{tabular}{@{}cc@{}}
\toprule
Name & \begin{tabular}{@{}cc@{}}Number of retrieved results \\ (this is likely to include duplicates)\end{tabular} \\ \midrule
DOE Data Explorer     & 41 results \\
Mendeley              & 83 results \\
Figshare              & 46 results \\
Zenodo                & 10 results \\
AmeriGEO              & 32 results \\
NIST                  & 35 results \\
Kaggle                & 6  results \\
DataCite              & 57 results \\
Google Dataset Search & 42 results \\ \bottomrule
\end{tabular}%
\end{table}

\subsection{Managing, Extracting and Drawing Conclusions from Dataset Searches}

In this subsection we present a combination of stages 5, 6 and 7 of the systematic review process namely that we screen the search output, extract data and trends and then draw conclusions around the topic of the review, to determine availability of open, annotated image datasets in AM processes.
We manually screened each retrieved dataset from each search service and after eliminating duplicate entries, our summary results are shown in Table~\ref{tab:screening_results}.
This shows that there are only a small number of datasets associated with additive manufacturing processes which are open discoverable using the 9 popular dataset search engines -- 10 datasets in total.

\begin{sidewaystable}[!htb]
\centering
\caption{Summary information on the results from screening}
\label{tab:screening_results}

\begin{tabular}{@{}lllllll@{}}
\toprule
  \textbf{\begin{tabular}[c]{@{}l@{}}Bibliographic \\ reference\end{tabular}} &
  \textbf{\begin{tabular}[c]{@{}l@{}}Technical \\ background\end{tabular}} &
  \textbf{Processing} &
  \textbf{Size} &
  \textbf{Format(s)} &
  \textbf{\begin{tabular}[c]{@{}l@{}}Type of\\  ML/DL\\ Applications\end{tabular}} &
  \textbf{\begin{tabular}[c]{@{}l@{}}Number \\ of citations\end{tabular}} \\ \midrule

\textbf{\cite{osti_1779073} }&
  \begin{tabular}[c]{@{}l@{}}Layer-wise powder \\ bed images \\ from 20 layers\end{tabular} &
  \begin{tabular}[c]{@{}l@{}}Labelled, \\ Manual\end{tabular} &
  \begin{tabular}[c]{@{}l@{}}140 images \\ and files of\\ annotations\end{tabular} &
  \begin{tabular}[c]{@{}l@{}}.tif\\ .npy\end{tabular} &
  Segmentation &
  5 \\
\textbf{\cite{osti_1896716}}&
  \begin{tabular}[c]{@{}l@{}}Layer-wise powder bed images \\ from two different powder bed\\  printing technologies.\end{tabular} &
  \begin{tabular}[c]{@{}l@{}}Labelled, \\ Manual\end{tabular} &
  \begin{tabular}[c]{@{}l@{}}6 datasets \\ and files of\\ annotations\end{tabular} &
  \begin{tabular}[c]{@{}l@{}}.tif\\ .npy\end{tabular} &
  Segmentation &
  4 \\
\textbf{\cite{osti_1923043}}&
  Later version of the dataset above. &
  \begin{tabular}[c]{@{}l@{}}Labelled, \\ Manual\end{tabular} &
  \begin{tabular}[c]{@{}l@{}}dataset\\  above\end{tabular} &
  \begin{tabular}[c]{@{}l@{}}.tif\\ .npy\end{tabular} &
  Segmentation &
  3 \\
\textbf{\cite{https://doi.org/10.17632/zyz6cznm5h.3}}&
  \begin{tabular}[c]{@{}l@{}}3D point cloud pre-processed to\\ achieve the target surface of the part, \\ then converted to a 2D depth \\ image.\end{tabular} &
  \begin{tabular}[c]{@{}l@{}}Labelled, \\ Manual\end{tabular} &
  43.4k images &
  .tif &
  Classification &
  no info \\
\textbf{\cite{https://doi.org/10.17632/8rm9d4ykbt.1}}&
  \begin{tabular}[c]{@{}l@{}}Feature-based CNN network and data for \\ laser powder bed fusion process.\end{tabular} &
  Labelled, &
  \begin{tabular}[c]{@{}l@{}}no images.\\ Converted\\ to .npy\end{tabular} &
  .npy &
  \begin{tabular}[c]{@{}l@{}}Training of \\ CNN\end{tabular} &
  no info \\
\textbf{\cite{https://doi.org/10.17632/h8tzpxkvdc.1}}&
  \begin{tabular}[c]{@{}l@{}}Using machine learning to predict \\ dimensions and qualify diverse part\\  designs across multiple additive\\  machines and materials.\end{tabular} &
  Labelled &
  4 images &
  \begin{tabular}[c]{@{}l@{}}.tif\\ .CSV\end{tabular} &
  Predictions &
  no info \\
\textbf{\cite{https://doi.org/10.17863/cam.84082}}&
  \begin{tabular}[c]{@{}l@{}}Images of the the extrusion\\ 3D printing process.\end{tabular} &
  Labelled &
  \begin{tabular}[c]{@{}l@{}}1,272,273\\  images\end{tabular} &
  \begin{tabular}[c]{@{}l@{}}.jpg\\ .CSV\end{tabular} &
  Error detection &
  no info \\
\textbf{\cite{LIU2022_normal}}&
  \begin{tabular}[c]{@{}l@{}}Emission images from in-situ monitoring\\  of additive manufacturing.\end{tabular} &
  \begin{tabular}[c]{@{}l@{}}Labelled, \\ Manual\end{tabular} &
  150 images &
  .png &
  \begin{tabular}[c]{@{}l@{}}Training,\\ Testing\end{tabular} &
  no info \\
\textbf{\cite{LIU2022_defect}}&
  \begin{tabular}[c]{@{}l@{}}Emission images from in-situ monitoring\\  of additive manufacturing.\end{tabular} &
  \begin{tabular}[c]{@{}l@{}}Labelled, \\ Manual\end{tabular} &
  150 images &
  .png &
  \begin{tabular}[c]{@{}l@{}}Training,\\ Testing\end{tabular} &
  no info \\
\textbf{\cite{Warren_2022}} &
  {\begin{tabular}[c]{@{}l@{}}Stainless Steel 316L printed \\ on an ExONE printer.\end{tabular}} &
  {Labelled} &
  336 images &
  .png &
  \begin{tabular}[c]{@{}l@{}}Detection, \\ Segmentation\end{tabular} &
  no info \\ \bottomrule
\end{tabular}%
\end{sidewaystable}

The results in Table~\ref{tab:screening_results} show that even for datasets gathered during the AM process  which use machine learning in some way, the related applications are very specific to particular AM sub-tasks or to the manufacture of specific AM outputs that may not be generalisable to other sub-processes or outputs. For example, the shapes of manufactured parts in the AM builds can be quite different, thus a dataset for detection and segmentation tasks for building gears, for example, may not be helpful or applicable when the object that needs to be justified changes significantly in shape to a type of valve, for example. 
%
%
The screening results which yielded these 10 datasets showed that the majority of them are used for detection and segmentation of large components in the manufacture rather than for detection of microstructures such as melt-pools with variable porosities which can be used in computer vision-based classification and detection tasks in AM.
Detecting microstructure defects in the melt-pool can be still done even when the geometry of the part changes for example from one gear part to a different gear part. Most of the datasets have been created and are openly available to support some other research question specific to the parts being manufactured rather than focusing on general issues such as porosity and microstructure faults as we do here. In fact, from these 10 screened results only \cite{https://doi.org/10.17632/zyz6cznm5h.3} and \cite{osti_1923043} provide detailed manuscripts which give a description the dataset and describe the value of the development of such a dataset to support microstructure detection in melt-pools. 


In addition to the two datasets previously mentioned, only one other dataset aligns with our selection criteria, specifically, the dataset provided in \cite{LIU2022_defect}. This comprises emission images originating from the melt-pools of additively manufactured Ti6Al4V. These findings show the limited availability of open and available image datasets from the AM domain that meet the requirements for use in machine learning applications, particularly in the context of microstructure detection. 

Even though we  searched and screened datasets from 9 individual databases, those searches may not have been exhaustive because the 9 databases that we used index data repositories only.  There is the possibility of  available datasets which are not indexed by the dataset search engines we used and are thus missing from our search. In order to investigate this as a possibility, we searched through the supplementary information in journal articles where research data is sometimes provided by authors.  For this we focused on Elsevier's journal ``Additive Manufacturing" \cite{Wicker2023} which is the top-ranked journal in the AM field. In the next section we describe our results when searching the archives of that journal.

\section{AM Datasets as Supplementary Information in Journal Articles}


The journal ``Additive Manufacturing''  is published by Elsevier and currently has 72 volumes. At the time of writing the latest issue is June 2023. This is a peer-reviewed journal that provides academic and world-leading industry researchers with high quality research papers and reviews in additive manufacturing. The scope of the journal comprises new technologies, processes, methods, materials, systems, and applications in AM and the journal's impact factor is 11.632 making it Q1.

The total number of articles included in the journal is 3,261 as of May 2023, and all those articles are searchable on the Scopus website. The searching and screening process, similar to the method used earlier, was repeated in order to discover articles which satisfy the dataset selection criteria described earlier though it is no longer necessary to specify the domain in the criteria since all articles are related to additive manufacturing. 

The journal ``Additive Manufacturing" is available on the ScienceDirect website of which the search engine supports Boolean operators and phrases. The initial searches were carried out on May 22nd 2023, using different combinations of keywords and operators in an iterative manner. When the search process was completed the results were up to  date as of May 24th, 2023. To ensure the search process is reproducible, the filter options as well as the final set of search queries and the associated number of relevant results retrieved per query with no filtering option selected, are shown below:



\begin{itemize}
\item ``machine learning”: 204 results
\item ``deep learning”: 68 results
\item ``machine learning” OR ``deep learning”: 217 results
\item image AND ``machine learning”: 177 results
\item image AND ``deep learning”: 64 results
\item image AND (``machine learning" OR ``deep learning"): 188 results
\end{itemize}
\noindent
These results indicate many papers published in this journal which address or use some form of machine learning in AM processes, but we are only interested in those papers which have included a freely available dataset as supplementary information to the paper so we applied the filter option that limits search results to only open access and open archive articles. Although many of the articles are not marked as open, the full text of the articles are still accessible via institutional subscriptions. Thus, we consider such articles count as open access and we leave the ``open access" criteria to be verified in the screening step where further insights will be given on the search results. Searches were then conducted without the ``open access" filter in order to retrieve as many potentially qualified results as possible. Then, the scope was narrowed down by applying additional terms such as ``machine learning”,``deep learning”, image and dataset with operators to limit the number of results and retrieve articles that are potentially relevant to the screening step.  The results of this are shown below.

\begin{itemize}
\item dataset AND ``machine learning": 93 results
\item dataset AND ``deep learning": 43 results
\item dataset AND (``machine learning" OR ``deep learning"): 96 results
\end{itemize}

We use the 96 articles from the journal search for further manual screening. While this set may contain  non-relevant results, we use these with the aim to not miss possible articles that can further support our review. 
The oldest of these articles was published in 2016 and the distribution of the number of retrieved articles across the years 2016 to 2023 is shown in Table~\ref{tab:search_journal}. The numbers in the table indicated that prior to the year  2020 there are only 7 research papers related to machine learning or deep learning in the Additive Manufacturing Journal which provide datasets. Since 2020, the number of articles about machine learning or deep learning increases each year. Due to the fact that volumes in 2023 are still in progress, there are only 11 results for 2023 in Table~\ref{tab:search_journal}.

\subsection{Managing, Extracting and Drawing Conclusions from Journal Searches}

\begin{table}[ht]
\centering
\caption{The numbers of articles in each year from journal searching}
\label{tab:search_journal}
\begin{tabular}{@{}cc@{}}
\toprule
\textbf{~~~~~~Year~~~~~~} & \textbf{\begin{tabular}[c]{@{}c@{}}Number of retrieved results \end{tabular}} \\ \midrule
2023     & 11 \\
2022     & 32 \\
2021     & 25 \\
2020     & 20 \\
2019     & 4 \\
2018     & 3 \\
2016     & 1 \\ \midrule
\textbf{Total}                 & 96 \\ \bottomrule
\end{tabular}%
\end{table}


Each of the 96 retrieved articles from the Additive Manufacturing journal were manually examined to ensure that information on their related datasets can be obtained from the articles. During this process, because of information on available dataset access can be hidden in certain parts of a paper such as sections that describe methods, results, discussions, conclusions and supplementary documents, each paper was examined until it was determined whether the associated dataset can, or cannot, be accessed. After the screening process we found that most journal articles focused on illustrating their methods, experiments and results, but only a minority of them clearly indicted how to access the datasets used in their research.

Of the total of 96 articles, the authors of 2 research articles declared that the datasets involved in their work are confidential or they do not have permission to publish their datasets. In another 13 research articles, the authors stated that datasets will only be available on request, which means these datasets are not openly available.
Of the remaining articles, 58 described their datasets or mentioned that datasets were used in their experiments but they did not indicate the way to access the related datasets. A further 8 articles are review papers with no specific dataset to be screened, and  there are 4 results are not relevant to the topic of this review, along with 1 duplicated result that has already been discovered earlier in Table~\ref{tab:screening_results}. 
This results in 86 articles being excluded for reasons of non-accessibility, non-relevance or duplication. The remaining 10 articles that provide links to their related datasets are further examined according to the criteria stated earlier and described in the next sub-section.      

The 10 journal articles were used for the final round of screening to decide whether their associated datasets meet the dataset selection criteria. Table~\ref{tab:screen_result_journal} presents a summary of the 10 results in the terms of technical background or topic, the way in which data is labelled (processed), dataset size, format, machine learning applications and the number of citations to the paper. The eligibility of each dataset was verified and analysed in terms of data types and characteristics making it suitable for the application of machine learning.

\begin{sidewaystable}
\centering 
\caption{Screening results and related information on the 10 papers with accessible datasets from the journal Additive Manufacturing.}
\label{tab:screen_result_journal}
\begin{tabular}{@{}llllllr@{}}
\toprule
Reference &
  \begin{tabular}[c]{@{}l@{}}Technical \\ background\end{tabular} &
  Data processing &
  Size &
  Format &
  \begin{tabular}[c]{@{}l@{}}Type of\\  ML/DL\\ Applications\end{tabular} &
  \multicolumn{1}{l}{\begin{tabular}[c]{@{}l@{}}Cited\\ by\end{tabular}} \\ \midrule

\textbf{\cite{LEE2020101444}} &
  \begin{tabular}[c]{@{}l@{}}Automated detection of part quality \\ during two-photon lithography \\ via deep learning.\end{tabular} &
  manully labeled &
  \begin{tabular}[c]{@{}l@{}}over \\ 5GB\end{tabular} &
  \begin{tabular}[c]{@{}l@{}}tiff, \\ .CSV, \\ npy\end{tabular} &
  Classification &
  16 \\
\textbf{\cite{WESTPHAL2022102535}} &
  \begin{tabular}[c]{@{}l@{}}Machine learning for the intelligent \\ analysis of 3D printing \\ conditions using environmental sensor.\end{tabular} &
  not image datasets &
  8MB &
  .CSV &
  Classification &
  8 \\
\textbf{\cite{AKBARI2022102817}} &
  \begin{tabular}[c]{@{}l@{}}MeltpoolNet: Melt-pool characteristic \\ prediction in Metal Additive Manufacturing \\ using machine learning.\end{tabular} &
  not image datasets &
  588kb &
  .CSV &
  Classification &
  8 \\
\textbf{\cite{ACKERMANN2023103585}} &
  \begin{tabular}[c]{@{}l@{}}Machine learning-based identification \\ of interpretable process-structure linkages \\ in metal additive manufacturing.\end{tabular} &
  not image datasets &
  1.93MB &
  .CSV &
  Regression &
  0 \\
\textbf{\cite{WESTPHAL2021101965}} &
  \begin{tabular}[c]{@{}l@{}}A machine learning method for defect detection \\ and visualization in selective laser sintering \\ based on convolutional neural networks.\end{tabular} &
  manully labeled &
  604MB &
  jpeg &
  Classification &
  55 \\
\textbf{\cite{KO2021101620}} &
  \begin{tabular}[c]{@{}l@{}}Machine learning and knowledge graph \\ based design rule construction for \\ additive manufacturing.\end{tabular} &
  not image datasets &
  \begin{tabular}[c]{@{}l@{}}25MB \\ (all files)\end{tabular} &
  x3p &
  \begin{tabular}[c]{@{}l@{}}Knowledge graph\\ based design.\end{tabular} &
  53 \\
\textbf{\cite{PAULSON2020101213}} &
  \begin{tabular}[c]{@{}l@{}}Correlations between thermal history \\ and keyhole porosity in laser powder \\ bed fusion.\end{tabular} &
  \begin{tabular}[c]{@{}l@{}}labeled with \\ parameters\end{tabular} &
  \begin{tabular}[c]{@{}l@{}}over \\ 10GB\end{tabular} &
  \begin{tabular}[c]{@{}l@{}}tiff, \\ .CSV, \\ xlsx\end{tabular} &
  \begin{tabular}[c]{@{}l@{}}Feature extraction,\\ regression\end{tabular} &
  52 \\
\textbf{\cite{OGOKE2022103250}} &
  \begin{tabular}[c]{@{}l@{}}Deep-learned generators of porosity \\ distributions produced during metal \\ Additive Manufacturing.\end{tabular} &
  \begin{tabular}[c]{@{}l@{}}generated data \\ by GAN\end{tabular} &
  678MB &
  tiff &
  \begin{tabular}[c]{@{}l@{}}Generative \\ Adversarial\\ Networks\end{tabular} &
  1 \\
\textbf{\cite{MOJUMDER2023103500}} &
  \begin{tabular}[c]{@{}l@{}}Linking process parameters with \\ lack-of-fusion porosity for laser powder \\ bed fusion metal additive manufacturing.\end{tabular} &
  not image datasets &
  16.4KB &
  .CSV &
  Regression &
  0 \\
\textbf{\cite{MENDOZAJIMENEZ2019100864}} &
  \begin{tabular}[c]{@{}l@{}}Parametric analysis to quantify process \\ input influence on the printed densities \\ of binder jetted alumina ceramics.\end{tabular} &
  not image datasets &
  11KB &
  xlsx &
  Regression &
  29 \\ \bottomrule
\end{tabular}%
\end{sidewaystable}

\begin{figure}[!ht]
    \centering
    \includegraphics[width=0.8\textwidth]{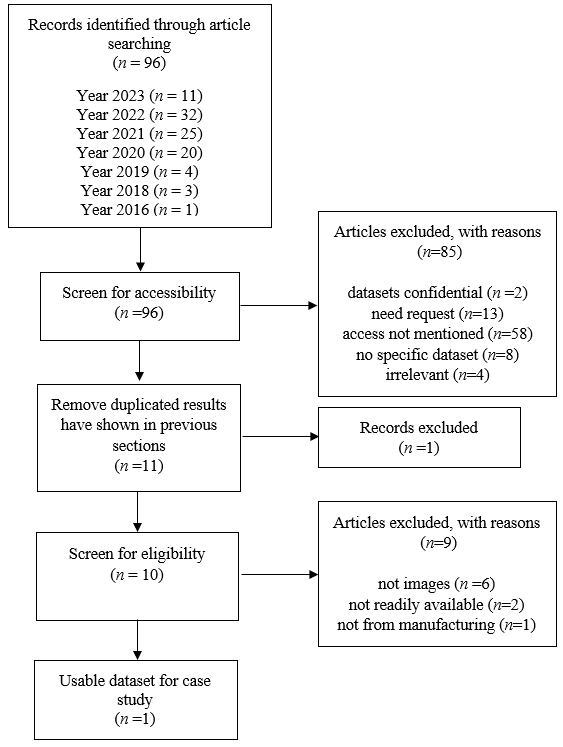}
    \caption{Overview of the screening process for articles in the Additive Manufacturing journal}
    \label{fig:AM_journal_screen}
\end{figure}

The full process of screening is illustrated in the flow diagram shown in Figure~\ref{fig:AM_journal_screen}. 
During the screening process it was discovered that 6 out of the 10 datasets are not image datasets, such as \cite{WESTPHAL2022102535} which uses numerical sensor   data that was gathered to record the manufacturing process; \cite{AKBARI2022102817} and \cite{KO2021101620} record the (numeric) parameters that describe the characteristic of the melt-pool; datasets descried in \cite{ACKERMANN2023103585}, \cite{MOJUMDER2023103500} and \cite{MENDOZAJIMENEZ2019100864} are series of processing parameters of the AM processes. The numbers and terms in these datasets are very specified to their related individual research topic. Because these 6 datasets do not meet the selection criteria that being in the format of images, they are excluded from our final result.

Of the remaining 4 datasets, they contain images and they are open  access, but not all are ready to be used by other research without further processing. For example, the dataset in \cite{LEE2020101444} is generated from two-photon lithography and the dataset from \cite{PAULSON2020101213} is based on X-ray and infrared imaging. The methods and experiments in these research papers are novel, but the data structures and the way of labelling are very specific to the associated research topic. Thus, these datasets are difficult to use generally and some of the labelling is hard to interpret without communication with the researchers in the original research team. For this reason, we consider these 2 image datasets are not within the selection criteria of readily available for general practice of machine learning research.

In 1 of the 2 remaining articles, the author of \cite{OGOKE2022103250} used Generative Adversarial Networks (GANs) \cite{goodfellow2014generative} to create stochastic realisations of synthetic parts from a limited dataset of experimental parts. This research can be considered as a good attempt to address the ``small data challenge". However the dataset that is published corresponding to this research only contains the GAN generated images as their experimental results. The ground truth image data that are used to develop their machine learning model, which we are really interested in, are not included. Furthermore, the GAN generated image datasets only include sample images of pores, resulting in only a single class in the dataset. Due to this, the use of the dataset is limited for general practice in machine learning.  For these reasons, we consider this GAN generated dataset in \cite{OGOKE2022103250}  is not suitable to be included in our further research in this thesis.

Finally, the  last article \cite{WESTPHAL2021101965} from screening the searching results proposed a machine learning method based on convolutional neural networks for defect detection in selective laser sintering, which is an additive manufacturing technology that uses a high-power laser to sinter powder of materials into a solid part \cite{wang20173d}.
The dataset involved in this research paper contains 4,000  images manually divided into 2 classes for defect detection tasks. The images in this dataset are also clearly separated into 3 categories for training, testing and validation and are ready to be used in the practice of machine learning by other researchers. In addition, there are total of 8,514 raw images available in the dataset for further potential use. This paper has been cited 55 times according to the Elsevier website, which is the highest number of citations among all papers in Table~\ref{tab:screen_result_journal}. This dataset of defect detection in AM shows very good potential that can be further unitised.

In summary, we originally targeted the identification of datasets which contain images of microstructures, such as in the melt-pool, from a monitored AM process and where these are annotated to highlight defects such as variable porosity or cracks. However, after scanning through all relevant articles in the journal, we find there are no image datasets of the melt-pool that are open access and readily available as supplementary information to articles from the journal. During the searching and screening process we discovered some articles, such as \cite{AKBARI2022102817} and \cite{KO2021101620} that described datasets related to the melt-pool, as well as \cite{MOJUMDER2023103500} that mentioned research related to lack-of-fusion porosity. However, these datasets contain process monitoring data in numerical format which record processing parameters and characteristics of melt-pools during the AM process. These datasets may offer support in some other machine learning or deep learning applications but are not suitable for computer vision based ML/DL approaches and tasks.


\section{Conclusions}


This systematic review of available datasets for vision-based defect detection in additive manufacturing processes comprised two primary investigations. The first investigation covered a comprehensive search across nine dataset databases, while the second focused on a highly-regarded Additive Manufacturing journal. The review reveals a significant gap in the availability of open and readily available image datasets  in the field of defect detection in additive manufacturing. We have conducted thorough searches through the dataset databases and through the journal archives  followed by manual examination and screenings of identified datasets, according to a set of selection criteria. 

It is essential to have appropriately annotated datasets for  effective application of machine learning, particularly computer vision-based deep learning in the context of additive manufacturing. Without such datasets we will not see  improvements in automatic defect detection in AM processes and the potential that this brings in terms of improved quality control and savings during manufacture when defects are detected and can be rectified or manufacturing of the remaining deposition layers prevented, will not be realised.

\newpage 

\section*{Appendix -- Search Processes from 9 Individual Dataset Search Services}
\label{apx:search}

We present the sequences of searches used when searching each of the dataset databases along with the numbers of datasets retrieved for each interactive search and a summary of the conclusions reached from the searches.
\begin{enumerate}
\item Database {\bf DOE Data Explorer}:
\begin{itemize}
\item “additive manufacturing” AND image: 41
\item additive AND manufacturing AND image: 41
\item additive manufacturing image: 41 
\item “additive manufacturing” AND image AND “machine learning”: 3 
\item “additive manufacturing” AND “machine learning”: 3
\item “additive manufacturing” AND image AND “deep learning”: 0
\item “additive manufacturing” AND “deep learning”: 0
\end{itemize}
Summary: the results from DOE Data Explorer are 41 results with 3 of them appearing to be duplicated datasets due to version difference. The resulting 38 datasets may not be in the additive manufacturing domain but this is something we examine in the screening phase of the systematic review process. 

\item Database {\bf Mendeley}:

Basic searches:
Filter: Data Types: Dataset
\begin{itemize}
\item additive manufacturing image: 40,299 (range is too big, too many non-relevant results)
\item additive AND manufacturing AND image: 1,008 (range is decreased but still many  non-relevant results)
\item “additive manufacturing” AND image: 176
\item “additive manufacturing” AND “machine learning”: 83
\item “additive manufacturing” AND image AND “machine learning”: 67
\item “additive manufacturing” AND image AND “deep learning”: 47
\item “additive manufacturing” AND image AND “machine learning” AND “deep learning”: 46 
\end{itemize}
Advanced search on Mendeley
Using keywords: any item with keywords “additive manufacturing” AND “machine learning” 
\begin{itemize}
\item KEYWORDS (“additive manufacturing” AND “machine learning”): 2 (the 2 results are relevant) 
\item KEYWORDS (“additive manufacturing” AND “deep learning”): 0
\end{itemize}

The results from basic searches on Mendeley  initially yield large numbers of retrieved datasets, mostly highly non-relevant. The scope was narrowed down by applying additional terms such as ``machine learning” and ``deep learning”, to limit the number of results. However, even doing so there are still a great proportion of the results that are non-relevant, Thus some advanced searches were carried out and 2 relevant results were obtained, but this scope may be too narrow for the current stage.    

Summary of results from Mendeley: we take the resulting number of 83 datasets which include the 2 datasets from the advanced search, though this output may contain many non-relevant results. 

\item Database {\bf Figshare}:

The basic search function on Figshare does not support Boolean operators.  For example, if ``AND” appears in the search string it will be treated as a search term. To utilise Boolean operators on Figshare, an advanced search must be conducted. After testing, it shows that the category filter for additive manufacturing on Figshare is case sensitive, so there are 3 terms (additive manufacturing, Additive manufacturing, Additive Manufacturing) to be selected in order to specify the search within the category of additive manufacturing. 

\begin{enumerate}
\item Basic search 1: 

Category: Deep learning, Item Type:dataset
\\Searching string: ``additive manufacturing": 2

This basic search is to retrieve datasets with the term ``additive manufacturing” from the Deep learning category. In fact, the only 2 results are the emission datasets from our own research, one is the emission image dataset of positive samples and the other is the dataset of negative samples. 

\item basic search 2:
Figshare does not provide a machine learning category, instead it gives 3 separated categories which are:

Category: Knowledge representation and machine learning, Item Type: dataset
\\Search string: additive manufacturing: 39 (results highly non-relevant)

Category: Adversarial machine learning; Item Type: dataset
\\Search string: additive manufacturing: 1

Category: machine learning not elsewhere classified; Item Type: dataset 
\\Search string: additive manufacturing: 6 

There are many non-relevant results from the 39+1+6 = 46 results due to the fact that the terms used are additive manufacturing which are regarded as 2 terms as there is no “” surrounding them to make them into a phrase to search on.

We then use a single term ``additive manufacturing" under the same filter conditions and obtain the following results.

Category: Knowledge representation and machine learning; Item Type: dataset
\\Search string: additive manufacturing: 0 

Category: Adversarial machine learning; Item Type: dataset
\\Search string: additive manufacturing: 0

Category: machine learning not elsewhere classified; Item Type: dataset  
\\Search string: additive manufacturing: 2

\item Basic search 3: filter: Category: additive manufacturing, Additive manufacturing, Additive Manufacturing; Item Type: dataset: 
\\Search string: 
\begin{itemize}
\item ``deep learning”: 3 
\item ``machine learning”: 5 
\end{itemize}
Even if the data type is not limited to images, the number of results shown are still small.

\item Basic search 4:
Item Type: dataset
\\Search string: additive manufacturing image: 270,179 

The results from this search are any item with additive or manufacturing or image. The number of results obtained is large but the contents are not relevant. This result is listed here only to record the search history and for investigation purposes.  

\item Advanced search on Figshare:
The aim of the advanced search is to use operators and syntax, such as AND and keywords, to precisely obtain datasets with desired terms and to fulfil the task that cannot be done using only a basic search.
The only filter used in this advanced search is Item Type: dataset
\begin{itemize}
\item :keyword:``additive manufacturing” AND :keyword:“machine learning”: 2
\item :keyword:``additive manufacturing” AND :keyword:“deep learning”: 3
\item :keyword:``additive manufacturing” AND :keyword:``deep learning” OR ``machine learning”: 3
\end{itemize}
3 results have been obtained that have the required keywords.  
\end{enumerate}

Summary of results from Figshare:  we may take as output, the resulting number of 46 datasets from the basic search 2.

\item From {\bf Zenodo}:
Basic search on Zenodo, filter: Type: {dataset}, Access Right: {open} 

The basic search was conducted using the following search strings:
\begin{itemize}
\item additive manufacturing image: 6,625
\item ``additive manufacturing” AND image: 10
\item additive AND manufacturing AND image: 4
\item ``additive manufacturing” AND ``machine learning”: 0
\item ``additive manufacturing” AND ``deep learning”: 0
\item ``additive manufacturing” AND image AND ``machine learning”: 0
\end{itemize}
The basic search on Zenodo supports search operators and keywords can be directly searched using strings and ``”. 

Summary of results from Zenodo: 10 datasets from the second search can be considered as an initial search output for further screening.

\item From {\bf AmeriGEO}:
Filters: None, but when the search is specified on datasets these are the results:
\begin{itemize}
\item ``additive manufacturing” AND image: 32
\item additive AND manufacturing AND image: 290 (there are too many non-relevant datasets in this)
\item additive manufacturing image: 6,410 (range is to big)
\item ``additive manufacturing” AND image AND “machine learning”: 0 
\item ``additive manufacturing” AND image AND “deep learning”: 0
\item ``additive manufacturing” AND image AND “machine learning” OR “deep learning”: 0 
\item ``additive manufacturing” AND image AND ``machine learning” AND “deep learning”: 0 
\item ``additive manufacturing” AND "machine learning”: 0
\end{itemize}
Summary of results from AmeriGEO: 32 datasets can be considered for screening.

\item From {\bf NIST} (with filter: manufacturing and dataset):
\begin{enumerate}
\item Basic search
Filter:Research Topic{Manufacturing}, Type{Dataset}
\begin{itemize}
\item additive manufacturing image: 77 (this range is too big)
\item ``additive manufacturing” image: 35 (reduced the range)
\item ``additive manufacturing” image ``machine learning”: 0
\item ``additive manufacturing” image ``machine learning” ``deep learning”: 0
\end{itemize}
Without filter:
\begin{itemize}
\item ``additive manufacturing” ``machine learning”: 0
\item ``additive manufacturing” ``deep learning”: 0
\end{itemize}
\item Advanced search on NIST: As the basic search on NIST does not support Boolean operators, to utilize Boolean operators on NIST, advanced search must be conducted.
\begin{itemize}
\item keyword=``additive manufacturing” AND keyword=``machine learning”: 0
\item keyword=``additive manufacturing” AND keyword=``deep learning”: 0
\end{itemize}
The results from NIST are stable in number when the terms are additive manufacturing and image, but when further limiting the application to “machine learning” it drops to 0. When using advanced search, it turns out that there is no item that carries the keywords ``additive manufacturing” and ``machine learning” or ``deep learning” at the same time.
\end{enumerate}
Summary of results from NIST: there are 35 results found in the advanced search on NIST. 

\item From {\bf Kaggle}:
There is no filter option, but searching is specified to a dataset type. The terms on Kaggle date search are treated as tags. It also seems to apply an AND logic operator between the tags by default. That is the reason why when more terms are added, the number of results are reduced.  
\begin{itemize}
\item ``additive manufacturing”: 6 (the page indicates 7 results, but in fact, only 6 results shown)
\item ``additive manufacturing” image: 3 (the page indicates 4 results, but in fact, only 3 results shown)
\item additive AND manufacturing AND image: 3 (same as above)
\item additive manufacturing image: 3 (same as above)
\item ``additive manufacturing” ``machine learning”: 0
\item ``additive manufacturing” AND image AND ``machine learning”: 0
\item ``additive manufacturing” AND image AND ``machine learning” AND ``deep learning”: 0
\end{itemize}
Overall, on Kaggle the results obtained from the search is much less than with Mendeley and Figshare. Even when just a single tag ``additive manufacturing” was used, the number of results obtained is still as small as 6 and the queries retrieved the same datasets so there were duplicates and overlaps in the results from individual queries. 

Summary of results from Kaggle: 6 datasets can be considered for screening

\item From {\bf DataCite}:
The filter applied in this search is Work Type: Dataset
\begin{itemize}
\item additive AND manufacturing AND image: 43
\item additive manufacturing image: 40
\item additive+manufacturing image: 40
\item ``additive manufacturing” AND image: 57
\item ``additive manufacturing”+image: 57
\item ``additive manufacturing” and image and ``machine learning”: 6
\item ``additive manufacturing” and image and ``machine learning” and ``deep learning”: 1 
\end{itemize}

Overall, as with previous search systems there are overlapping datasets among the results from different queries. 

Summary of results from DataCite: 57 datasets can be considered for screening

\item From {\bf Google Dataset Search}:
\begin{itemize}
\item additive and manufacturing and computer vision: 10
\item ``additive manufacturing” and image: 25
\item additive and manufacturing and image: 32
\item additive manufacturing image: 42
\end{itemize}
Summary of results from Google Dataset Search: 42 unique and non-duplicated datasets will be input into the next stage
\end{enumerate}


\printbibliography

\vspace{1cm}

{\includegraphics[width=1in,height=1.25in,clip,keepaspectratio]{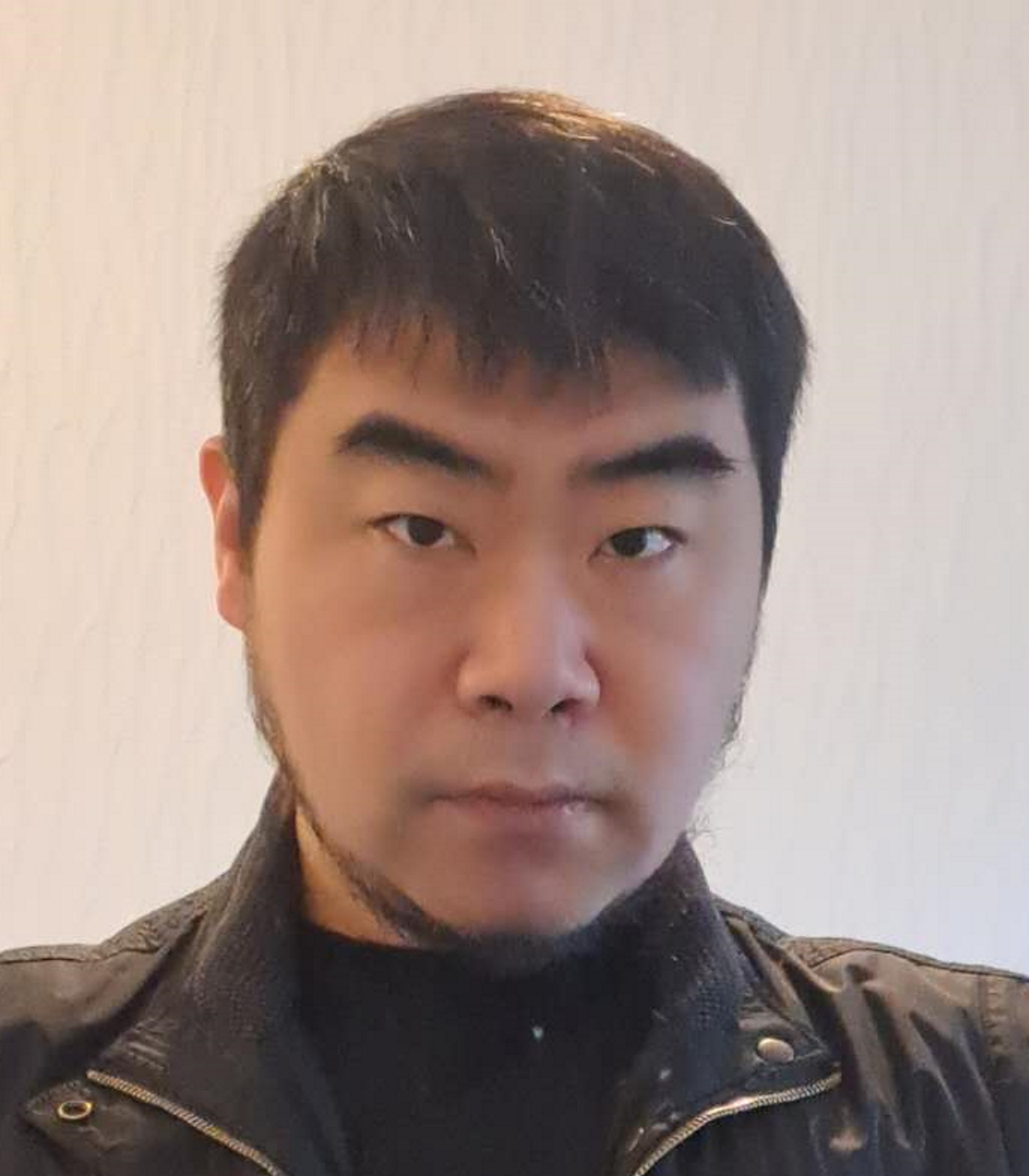}}
{Xiao Liu} received his B.S. degree in Mechatronic Engineering and M.S. degree in Electronic Engineering from Dublin City University, Ireland, in 2010 and 2012, respectively. He served as a Research Assistant in the School of Mechanical and Manufacturing Engineering at Dublin City University from 2013 to 2017. Since 2018, he has been pursuing his Ph.D. in the School of Computing Science at Dublin City University. His research interests encompass natural language processing and additive manufacturing.

\vspace{1cm}

{\includegraphics[width=1in,height=1.25in,clip,keepaspectratio]{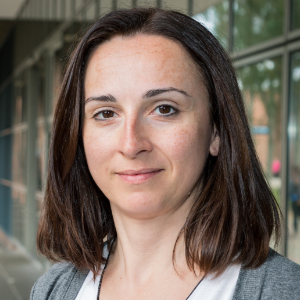}}
{Alessandra Mileo FHEA} holds a PhD (2006) in Computer Science from the University of Milan, Italy. She is an Associate Professor in the School of Computing at Dublin City University, a Principal Investigator at  the INSIGHT Centre for Data Analytics and a Funded Investigator at the I-Form Centre for Advanced Manufacturing. She secured over 1 million euros in national, international and industry-funded projects, publishing 90+ papers in the area of knowledge representation and reasoning, semantic technologies, knowledge graphs, stream processing, neuro-symbolic computing and explainable AI.
Dr. Mileo is a member of The Association for the Advancement of Artificial Intelligence (AAAI), the Italian Association for Artificial Intelligence (AI*IA), the Association for Logic Programmiing (ALP), and Steering Committee member of the Reasoning Web and Rule Systems association (RRA), as well as Editorial Board member of the Neuro-Symbolic AI Journal, IOS Press.

\vspace{1cm}

{\includegraphics[width=1in,height=1.25in,clip,keepaspectratio]{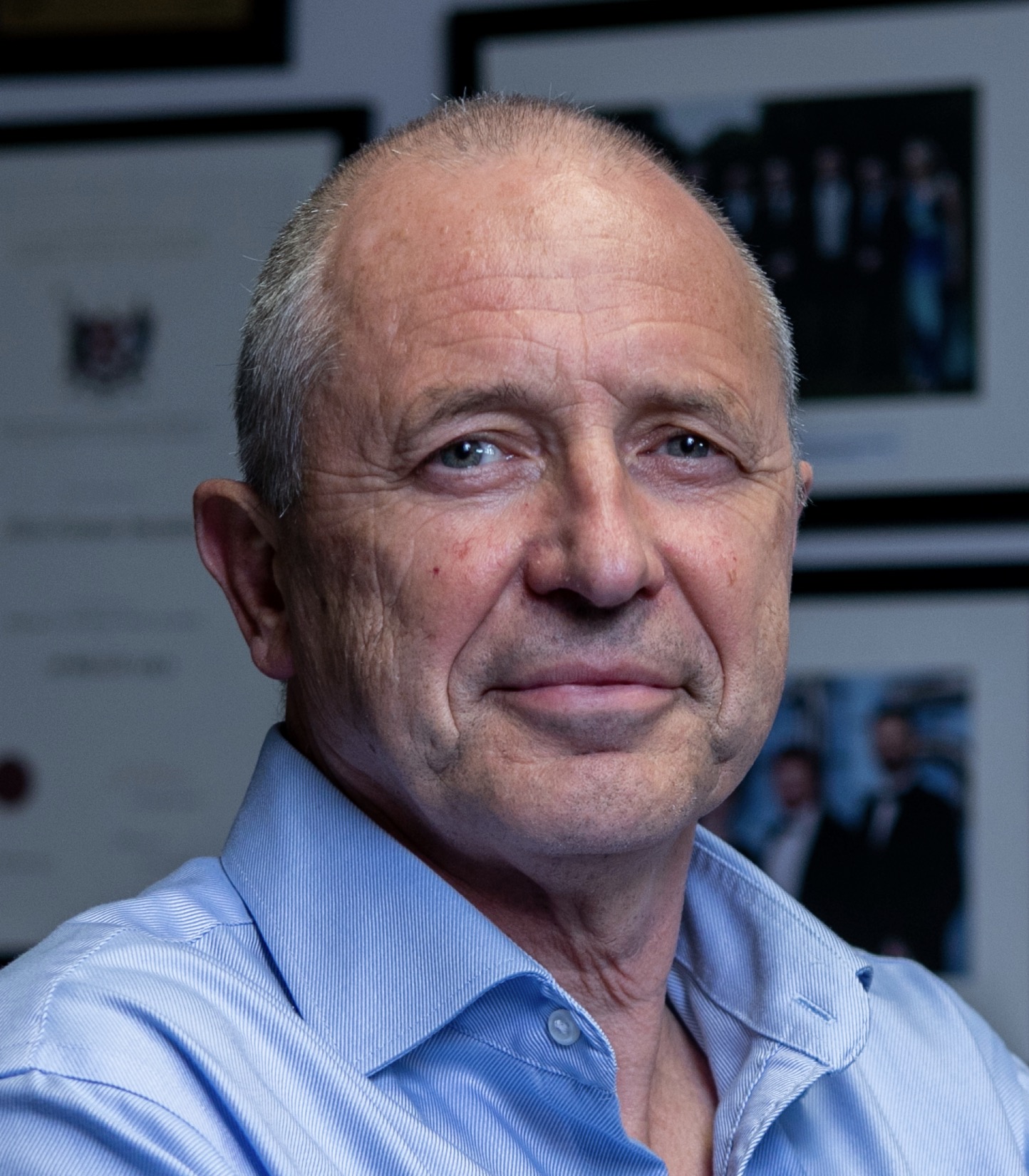}}
{Alan F. Smeaton MRIA FIEEE PFHEA FAAIA} (M'87--SM'12--F'16) 
Alan Smeaton was born in Dublin, Ireland and holds the B.Sc. (1980), M.Sc. (1982) and PhD (1987) degrees in computer science from University College Dublin. His major research interest is in helping finding people to find information, and trying to discover why they need that information and if it is information they previously had, why they have forgotten it.

Since 1987 he has been on the Faculty at Dublin City University where he has previously served as Head of the School of Computing (twice) and Dean of Faculty. He is a founding Director of the Insight Centre for Data Analytics, one of the largest publicly-funded research centres in Europe, and he was appointed Professor of Computing in 1997. 

Prof. Smeaton is an elected member of the Royal Irish Academy and winner of the Academy’s Gold Medal in Engineering Sciences in 2016. He is a member of the ACM and winner of the 2022 ACM SIGMM (Special Interest Group in Multimedia) Award for Outstanding Technical Contributions to Multimedia Computing, Communications and Applications.  He is the author of more than 700 research papers and book chapters with more than 22,000 citations and he has an h-index of 72.


\end{document}